\documentclass[preprint,review,12pt]{elsarticle}

\usepackage{amssymb}

\usepackage{booktabs}
\usepackage{amsmath,amssymb,amsfonts}
\usepackage{algorithmic}
\usepackage{graphicx}
\usepackage{textcomp}
\usepackage{hyperref}
\usepackage{multirow}
\usepackage[]{footmisc}
\usepackage{lscape}
\usepackage[table,xcdraw]{xcolor}
\graphicspath{{.}{./Figure/}}

\journal{Natural Language Processing}

\begin{document}

\begin{frontmatter}



\title{Sentiment Analysis and Opinion Mining on Educational Data: A Survey}


\author[1]{Thanveer Shaik\corref{cor1}}
\ead{Thanveer.Shaik@usq.edu.au}


\address[1]{School on Mathematics, Physics, and Computing, University of Southern Queensland, Toowoomba 4350, Australia}

\author[1]{Xiaohui Tao\corref{cor2}}
\ead{Xiaohui.Tao@usq.edu.au}


\author[2]{Christopher Dann}

\ead{Chris.Dann@usq.edu.au}


\address[2]{School of Education, University of Southern Queensland, Toowoomba 4350, Australia}

\author[3]{Haoran Xie}
\ead{hrxie@ln.edu.hk}

\author[1]{Yan Li}
\ead{Yan.Li@usq.edu.au}

\address[3]{Department of Computing and Decision Sciences, Lingnan University, Tuen Mun, Hong Kong}

\author[1]{Linda Galligan}
\ead{Linda.Galligan@usq.edu.au}


\address[4]{School of Business, University of Southern Queensland, Springfield 4300, Australia}

\begin{abstract}

Sentiment analysis AKA opinion mining is one of the most widely used NLP applications to identify human intentions from their reviews. In the education sector, opinion mining is used to listen to student opinions and enhance their learning-teaching practices pedagogically. With advancements in sentiment annotation techniques and AI methodologies, student comments can be labelled with their sentiment orientation without much human intervention. In this review article, 1) we consider the role of emotional analysis in education from four levels: document level, sentence level, entity level, and aspect level, 2)sentiment annotation techniques including lexicon-based and corpus-based approaches for unsupervised annotations are explored, 3)the role of AI in sentiment analysis with methodologies like machine learning, deep learning, and transformers are discussed, 4) the impact of sentiment analysis on educational procedures to enhance pedagogy, decision-making, and evaluation are presented. Educational institutions have been widely invested to build sentiment analysis tools and process their student feedback to draw their opinions and insights. Applications built on sentiment analysis of student feedback are reviewed in this study. Challenges in sentiment analysis like multi-polarity, polysemous, negation words, and opinion spam detection are explored and their trends in the research space are discussed. The future directions of sentiment analysis in education are discussed.    

\end{abstract}




\end{frontmatter}


\section{Introduction}\label{sec:Introduction}

Sentiment analysis, also known as opinion mining is a natural language processing technique (NLP) technique to identify the emotional tone behind a text~\cite{Bansal2022}. The technique has been widely used to extract user opinions on products and services from their reviews and create actionable knowledge for an entity~\cite{Ligthart2021}. This enables businesses to improve their strategies and gain insights into customers' feedback about their products and services. Opinion mining is a multi-disciplinary field that includes machine learning, NLP, sociology, and psychology to detect underlying customer or user opinions. Social media sites like Twitter, Facebook, and Instagram are significant sources of user opinions. The analysis of these sources to extract sentiment or opinions of users started a decade ago~\cite{Tao2016, Zhou2013}. 

Advancements in technology have transformed fields like healthcare~\cite{Tao2021, Shaik2022Fed} and education by adopting AI and NLP techniques. In the education domain, student feedback plays a vital role in evaluating and analysing learning management systems, teaching, pedagogical procedures, and courses~\cite{Elfeky2020}. Educational institutions use student feedback surveys at the end of each semester to record their opinions on courses enrolled in the semester~\cite{McKinney2022}. The feedback comprises both qualitative and quantitative data which includes demographics of students, courses, ratings, and comments. The quantitative data can provide a statistical understanding of feedback on the courses, but the students' intent can be determined based on qualitative data analysis. The textual comments have to be preprocessed through NLP techniques such as feature extraction, and feature selection for further analysis~\cite{Zhao2021}. Qualitative data analysis enables listening to students' opinions on each course, content, and teaching.

In sentiment analysis, the initial step is to label text with emotional tags like positive, negative, or neutral which denotes students' emotional opinions on the services provided. The levels of the sentiment analysis differ based on application requirements~\cite{Zhang2021}. Few applications might need an overview of the student satisfaction report and few might need fine-grained analysis at the topic level to understand which aspect of the course delivery has negative reviews for improvisation. The manual annotation or labelling of the sentiment orientation is time-consuming~\cite{Liu2019} and requires many resources with pedagogical understanding in education. This challenge has been addressed by developing different sentiment annotation approaches using lexicons and corpus. These techniques act as unsupervised techniques for initial level understanding of the student feedback. AI's role in sentiment analysis is unavoidable as it assists to process and analysing a large number of student comments~\cite{Zhu2021}. AI methodologies like machine learning, deep learning, and transformers~\cite{Acheampong2021} are capable of learning student opinions with attention mechanisms and classifying or predicting their emotions for unlabelled student comments~\cite{Kuleto2021}. The unsupervised sentiment annotation techniques and the AI methodologies overcome the challenge of manual labelling to a certain extent. In this review article, the following research questions were explored:

\begin{itemize}
    \item What is the role of sentiment analysis and opinion mining in education?
    \item What are the different levels of sentiment analysis in the education domain?
    \item What are the current trends in sentiment annotation of educational data? 
    \item What are the challenges of adapting sentiment analysis in the education domain?
\end{itemize}

This study explores sentiment analysis's role in enhancing educational procedures, different levels of sentiment analysis, and sentiment annotation techniques. The challenges that affect sentiment analysis in the education domain are discussed in detail. The contributions of this research are as follows:

\begin{itemize}
    \item Enhancement in educational domain procedures by adopting the sentiment analysis.  
    \item Overview on sentiment annotation techniques and role AI in sentiment analysis.
    \item Exploring trends and challenges in sentiment analysis that need to be addressed to be adopted in the education domain.
\end{itemize}

The remainder of the paper is organised into sections. Different levels of sentiment analysis adopted in education have been explored in Section~\ref{levels}. Sentiment annotation techniques and AI methodologies involved in sentiment analysis are reviewed in Section~\ref{annotations}. Section~\ref{impact} discusses the impact of sentiment analysis on educational procedures like evaluation, decision-making, and pedagogical concepts. Section~\ref{challenges} presents the challenges in sentiment analysis and its trends. In Section~\ref{future}, the future directions of sentiment analysis in the educational domain are discussed. Finally, the article concludes in Section~\ref{Conclusions}.

\section{Sentiment Analysis in Education}\label{levels}

\begin{figure}
    \centering
    \includegraphics[width=50mm]{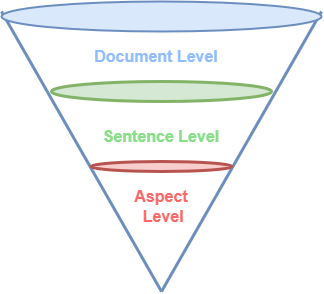}
    \caption{Sentiment Analysis in Education}
    \label{fig:inverted_pyramid}
\end{figure}
Sentiment analysis has the potential to extract student opinions at the document level, sentence level, entity level, and aspect level with their sentiment orientation~\cite{Ligthart2021}. The document level analyses a comment and determines the overall sentiment of the comments towards a course is positive, negative, or neutral. At the sentence level, the sentiment extracts from each sentence and helps calculate a course's positives and negatives. Entity-level sentiment extraction combines entity and sentiment analysis to provide student opinion on an entity like tutor, course, and assignment. Aspect-based sentiment analysis is a fine-grained analysis of different data categories in a comment and identifies sentiment orientation on each data category. According to an education application, student feedback data will be analysed at different levels. For example, a decision-making application would consider document-level sentiment analysis, and to understand student engagement, the analysis would be at the aspect level.

\subsection{Document Level}
In the education domain, developing sentiment extraction tools would require research resources. Hence, many institutions employ general sentiment extraction tools which are not domain specific. Dolianiti et al.~\cite{DOLIANITI2019} compared the performance of five commercial sentiment analysis tools IBM Watson Natural Language Understanding, Microsoft Azure Text Analytics API, OpinionFinder 2.0, Repustate, and Sentistrength with educational domain-specific tools in the document and sentence levels. In that study, two educational datasets were used which contain student forum posts of two courses across a semester in the learning management system. The sentiment orientation of the forum posts in datasets was manually annotated in two different versions: document level and sentence level. Four education domain tools, two for each course, were developed using SVM and k-fold cross-validation techniques. The study reported that educational-domain tools outperformed commercial tools in one of the courses at both document and sentence levels. Faculty performance evaluation using document-level sentiment analysis was discussed by Ahmed et al.~\cite{ahmad2018sentiment}. Two machine learning classifiers SVM and NB were trained on a preprocessed dataset of 5000 comments and achieved an accuracy of 72.80\% and 81\% respectively. 

\subsection{Sentence Level}
Sivakumar et al.~\cite{Sivakumar2017} used the cosine similarity method to measure semantic relatedness between aspect words and student opinion sentences. The dataset used in that study was extracted from Twitter API, preprocessed, and classified the comments into seven aspects at the sentence level. Three machine learning algorithms decision trees, SVM, and NB were used to classify the sentences into different aspects. Subjective sentences were extracted using parts-of-speech(POS) tagging and their sentiment orientation was attributed using a lexicon-based approach SentiWordNet. Nikoli et al.~\cite{Nikoli2020} conducted a sentence-level analysis to extract one or more aspects in the sentences and classify their polarity into positive or negative sentiment. The authors used SVM, cascade classifier, and rule-based methods for aspect extraction and merged them. Similarly, the sentiment was detected using a merged component of SVM and a dictionary-based approach. Overall, the research was able to detect negative sentiment with an F-measure of 0.94 whereas positive sentiment was identified with an F-measure of 0.83 only. 

\subsection{Entity Level}
In an educational context, student feedback on entities such as teachers, learning management systems, or a specific concept of course content would help analyse their opinions. Yang et al.~\cite{Yang2021} discussed automatic entity extraction in educational contexts such as content, teacher, lesson, and curriculum. Named Entity Recognition (NER) acts an important role in entity-level sentiment analysis which could a word or phrase that clearly detects a person, company or location~\cite{Li2022tkde}. Ding et al.~\cite{Ding2018} developed an entity-level sentiment analysis tool, SentiSW, to extract sentiment and entity from comments in the form of the tuple "(sentiment, entity)". The authors used TF-IDF and Doc2Vec to evaluate sentiment classification and achieved the best performance with Gradient Boosting Tree and Linear Support Vector machine. The results show that the proposed SentiSW tool outperformed existing tools like SentiStrength and SentiStrength-SE in positive and neutral comments classification. Li et al.~\cite{Li2017} proposed a novel approach to learning latent sentiment scopes at the entity level with named entities and sentiment is highlighted. The authors used graphical models to form nine different types of nodes at each word's position to determine before entity B, entity E, and after entity A. Based on their polarity, each word is annotated as positive, negative or neutral sentiment. 

\subsection{Aspect Level}
Aspect-based sentiment analysis (ABSA) is the most widely adopted approach for sentiment extraction in education. This approach provides a fine-grained analysis of educational data at phrase or sentence levels and extracts opinions or emotions at key aspects or entities. To perform ABSA, sentiment analysis techniques have to be ensembled with topic modelling techniques such as LDA, Latent Semantic Analysis (LSA), Non-negative matrix factorisation (NMF), and Probabilistic latent semantic analysis (PLSA). A detailed definition and applications of the topic modelling techniques in educational data were discussed by Shaik et al.~\cite{Shaik2022}.

Rosalind et al.~\cite{MelbaRosalind2022} proposed an ABSA system to extract student satisfaction on online courses in Coursera using a machine learning algorithm. Aspects in the student reviews were retrieved using unsupervised and semi-supervised LDA techniques. The actual reviews were segmented into sentences and then the aspects and their sentiment polarity were estimated. A customised lexicon was used to calculate the sentiment polarity of the sentences and it yield positive polarity for learning aspects, lab, job/career, and grade/test, and negative polarity for instructor, content, course, fee, and teaching. The maximum entropy classifier was trained and tested for the classification of multi-aspects and sentiments. The proposed model achieved an accuracy of 80.67\% for aspect-based sentiment classification. Similarly, Edalati et al.~\cite{Edalati2021} conducted aspect-based opinion mining on student comments from the Coursera platform to study the experience of conducting lectures and taking classes online. The authors deployed RF, SVM, decision tree, and deep learning models to identify teaching related-aspects and predict student opinions on the aspects. RF has the best performance with an F1 score of 98.01\% and 99.43\%  in aspect identification and aspect sentiment classification respectively. 

Wehbe et al.~\cite{Wehbe2021} proposed a Spatiotemporal sentiment analysis framework to analyse an education-related Twitter dataset using ASBA, sentiment analysis, and emotional analysis based on specific time and location. This framework used five machine learning classifiers decision tree, RF, multinominal, SVC, and gradient boosting to classify six emotions such as happiness, surprise, sad, anger, fear, and disgust, and four aspects educational rights, job security, financial security, safety, and death. In both aspects and emotions classification, RF had the best performance with an accuracy of 96.99\% and 88.72\% respectively. Aspect-level sentiment analysis can be applied to verbal speech to extract student emotions and predict their performance in collaborative learning. Dehbozorgi et al.~\cite{Dehbozorgi2021} proposed a multi-class emotion analysis on students' speech in teams and their performance. The authors classified emotions such as anger, happiness, sadness, surprise, and fear using Text2Emotion, a python package. Rule-based POS tagging was used to extract aspects, and then the k-nearest neighbour (KNN) algorithm was used to predict student performance by connecting extracted aspects and emotions. Kastrati et al.~\cite{Kastrati2020} proposed an aspect-level sentiment analysis framework to identify sentiment or opinion polarity towards a given aspect related to MOOCs. The proposed framework used weakly supervised annotation to identify aspect categories in unlabelled students' reviews. This reduces the need for manually annotated datasets in deep learning techniques. The authors used the convolutional neural networks (CNN) model for the prediction of aspects and sentiment classification. The proposed framework achieved an F1 score of 86.13\% for aspect category identification and 82.10\% for aspect sentiment classification. The research works adopted AI in different levels of sentiment analysis are presented in Tab.~\ref{tab:SA_Levels}.

\begin{table*}[!h]
\scriptsize
\caption{Sentiment Analysis in Education}
\label{tab:SA_Levels}
\resizebox{\textwidth}{!}{%
\begin{tabular}{@{}llll@{}}
\toprule
SA Level &
  Algorithms &
  Application &
  References \\ \midrule
\multicolumn{1}{l}{Document Level} &
  \multicolumn{1}{l}{SVM, NB} &
  \multicolumn{1}{l}{\begin{tabular}[c]{@{}l@{}}Evaluate the documents' overall opinion \\ towards a context is positive or negative.\end{tabular}} &
  \multicolumn{1}{l}{\cite{DOLIANITI2019,ahmad2018sentiment}} \\ \midrule
\multicolumn{1}{l}{\cellcolor[HTML]{FFFFFF}Sentence Level} &
  \multicolumn{1}{l}{\begin{tabular}[c]{@{}l@{}}SVM, NB, \\ cascade classifier, \\ rule-based methods\end{tabular}} &
  \multicolumn{1}{l}{\begin{tabular}[c]{@{}l@{}}Breakdown educational feedback and perform \\ sentiment analysis at each sentence in a document\end{tabular}} &
  \multicolumn{1}{l}{\cite{Sivakumar2017,Nikoli2020}} \\ \midrule
\multicolumn{1}{l}{\cellcolor[HTML]{FFFFFF}Entity Level} &
  \multicolumn{1}{l}{\begin{tabular}[c]{@{}l@{}}Gradient boosting tree, \\ SVM, NER, TF-IDF, \\ Doc2Vec\end{tabular}} &
  \multicolumn{1}{l}{\begin{tabular}[c]{@{}l@{}}Analyse sentiment or opinion in feedback \\ towards an entity in educational domain\end{tabular}} &
  \multicolumn{1}{l}{\cite{Yang2021,Li2022tkde,Ding2018,Li2017}} \\ \midrule
\cellcolor[HTML]{FFFFFF}Aspect Level &
  \begin{tabular}[c]{@{}l@{}}LDA, RF, SVM, \\ decision tree, \\ CNN, KNN\end{tabular} &
  \begin{tabular}[c]{@{}l@{}}Understand positive or negative \\ aspects in educational practices.\end{tabular} &
  \cite{MelbaRosalind2022,Edalati2021,Wehbe2021,Dehbozorgi2021,Kastrati2020} \\ \bottomrule
\end{tabular}}
\end{table*}

Sentiment analysis can draw student opinions on different levels of feedback and provide insights to educational institutions for making informed decisions. However, the sentiment analysis can be performed only after labelling or annotating the text data with its sentiment orientation such as positive, negative, or neutral. There are supervised and unsupervised approaches to labelling or annotating student feedback. In the next section, sentiment annotation techniques that are adopted in education are discussed.

\section{Sentiment Annotation techniques in Education}\label{annotations}
Sentiment annotation is to label a document, sentence, or phrase with its semantic emotions that could be positive, negative, or neutral. The annotation can be fine-grained with micro-level analysis to extract precise emotions of users towards a service or product. This annotation mechanism can be applied to educational applications where student feedback towards learning management system, course content, learning-teaching practices, and instructor teaching abilities. Based on student feedback, the educational infrastructure can be streamlined, can predict student performance, and support students with personalised learning. However, an increase in the number of student enrolments and their feedback made the manual sentiment annotation process next to impossible as it requires massive resources and time. Advancements in Artificial Intelligence and NLP methodologies led to a variety of tools for sentiment analysis. The tools are built on a lexicon-based approach, corpus-based approach, machine learning, deep learning, and transformer approaches.

Yeruva et al.~\cite{Yeruva2020} discussed differences between human annotators and machine annotators in sentiment analysis tasks. The authors presented a human-in-the-loop approach to explore human-machine collaboration for sentiment analysis. The sentiment annotation data obtained from 60 human annotators were compared with machine annotations extracted from six toolkits CoreNLP's sentiment annotator, Vader, TextBlob, Glove+LSTM, LIME, and RoBERTa. The authors conducted coefficient correlation analysis to understand the importance of features, words, and topics in human-machine collaboration. The results show that computational sentiment analysis has high performance and can be applied to text analysis.

\subsection{Unsupervsied Annotation Techniques}
\subsubsection{Lexicon-based Approach}
Sentiment and opinion words often act a vital role in the sentiment annotation of a document or sentence~\cite{Catelli2022}. Identifying the sentiment and opinion words can help to categorise the sentiment in an unsupervised manner~\cite{DOLIANITI2019}. In a lexicon-based approach, a sentiment dictionary with lexical units like words or phrases and their corresponding sentiment orientation like real values (eg: ranging from -1 to +1), classes (eg: positive, negative, or neutral), fine-grained classes like (eg: very positive to very negative). The sentiment orientation is based on the polarity of the content words like adjectives~\cite{Hatzivassiloglou1997,taboada2006sentiment}, adverbs~\cite{benamara2007sentiment}, verbs~\cite{vermeij2005orientation}, nouns~\cite{neviarouskaya2009compositionality} and phrases in a sentence or document. Different lexicon-based approaches are developed for the English language based on the core idea of the sentiment dictionary. The approaches are SentiWordNet~\cite{SentiWordNet}, Opinion Finder~\cite{wilson2005opinionfinder}, Bing Liu's Opinion Lexicon~\cite{Liu2012}, MPQA subjectivity lexicon~\cite{wilson2005recognizing}, Harvard General Inquirer~\cite{stone1966general}, AFINN~\cite{nielsen2011new}, SentiFul~\cite{neviarouskaya2009sentiful}, Vader~\cite{Eytan2016}, TextBlob~\cite{loria2018textblob} and so on. 

Tzacheva et al.~\cite{Tzacheva2021} used National Research Council (NRC)~\cite{ortony1990s,mohammad2013crowdsourcing} lexicon to label student feedback with fine-grained emotions such as joy, fear, trust, anger, sadness, disgust, and anticipation for teaching innovation assessment. The authors evaluated the impact of active learning methodologies like flipped classroom approach~\cite{Maher2015} and lightweight teams~\cite{Latulipe2015, MacNeil2016} implementation using emotion detection and sentiment analysis. The results show that the trust component increased from the time period of 2015 to 2020. Existing lexicon methods can be improved by modifying the strategies adopted to develop the method. Rosalind et al.~\cite{MelbaRosalind2022} proposed steps to improve the existing Bing lexicon and develop a customised sentiment lexicon (CSL). The authors calculated sentiment polarity on academic course feedback texts using Bing sentiment Lexicon and CSL approaches. The process is to tokenise a sentence using the bag-of-words (BoW) method and calculate the polarity score of each word with Bing and CSL methods to output the cumulative polarity score of the sentence. The estimated cumulative score decides the sentiment orientation of the sentence. The CSL approach performed better than the existing Bing sentiment lexicon at the document level, detection  of opinion words, and polarity scoring. 

Lexicon-based approaches can perform unsupervised labelling of educational data and avoid manual labelling. The challenging part of the lexicon-based approach in sentiment analysis is domain or context understanding. The dictionary-based approach has a simple mapping of certain keywords with their sentiment. The words outside the listed keywords or opinions of students in the context are ignored. To avoid this, corpus-based sentiment analysis can be adopted. 

\subsubsection{Corpus-based Approach}
A corpus is a collection of texts on a specific topic or domain. The corpus-based approach in sentiment analysis is based on con-occurrence statistics and syntactic patterns of words in text corpora. This helps to enhance the sentiment lexicons with prior information about words across the semantic orientation of sentiment. The underlying intuition of corpus-based techniques for sentiment lexicon is to calculate the semantic distance between a word and a set of positive or negative words to estimate the semantic polarity of the target word. This process could help to adapt the domain-independent sentiment lexicon to a domain-specific lexicon~\cite{Alqasemi2017}.

\subsection{Supervised Annotation Techniques}
The lexicon and corpus-based approaches are capable to label text data in student feedback with sentiment orientation. However, processing and analysing the opinions and emotions of multiple students' feedback or any other feedback would be a challenging task. AI methods have the potential to get trained on preprocessed NLP data and be able to classify or predict the sentiment orientation of the data. 
\subsubsection{Machine Learning}
A learning management system (LMS) plays an important role to provide access to course content for offline and online students and record their engagement. Osmanoglu et al.~\cite{Osmanolu2020} analysed student feedback gathered from a university to assess the course materials. The authors used machine learning techniques to classify the materials into positive, negative, or neutral sentiments and then improve the course materials with negative feedback for their upcoming semester. Six classifiers multinomial logistic regression, decision tree, multi-layer perceptron, XGBoost, support vector classifier, gaussian Naive Bayes, and k-nearest neighbours were used after preprocessing the student comments. Logistic regression was able to perform better than the other five classifiers. To implement a feedback analysis system, Lwin et al.~\cite{Lwin2020} processed student textual comments along with quantitative ratings. The quantitative rating scores were clustered using the K-means clustering algorithm. The authors also used six classifiers support vector classifiers, logistic regression, multi-layer perceptron, and random forest to classify the clustered dataset. The textual comments were manually labelled as positive or negative as their sentiment. A naïve Bayes classifier was used to train the labelled dataset to classify the comments into positive or negative. Faizi~\cite{Faizi2022} proposed a machine learning approach to classify learners' feedback sentiment towards Youtube educational videos. Traditional machine learning methods such as random forest, logistic regression, Naïve Bayes, and SVM were adopted to learn the feedback datasets and classify the learner comments with sentiment as positive or negative. The SVM algorithm was able to perform better than other models with an accuracy of 92.82\% on a combination of unigrams and bigrams and an accuracy of 92.67\% for associations of unigrams and trigrams. The algorithms discussed so far might have certain limitations individually. Kaur et al.~\cite{Kaur2022} proposed a hybrid framework based on three classifiers random forest, logistic regression, and SVM models. The authors used the lexicon-based method sentiwordnet to label student comments with positive or negative sentiments. The proposed hybrid classifier was trained on the labelled dataset and compared its performance with the SVM model at different test-training ratios. The hybrid classifier outperformed the SVM model in all classification metrics.  

\subsubsection{Deep Learning}
Early identification of students who are likely to fail a course can be predicted using sentiment analysis and deep learning techniques. Yu et al.~\cite{Yu2018} proposed a deep learning model convolutional neural networks (CNN) to learn structured data like attendance, grades and unstructured text feedback from 181 undergraduate students. The text feedback was manually annotated with positive, negative, or labelling based on the Self-Assessment Manikin rating scale~\cite{lang1980behavioral}. The authors trained support vector machines and CNN models with structured and unstructured feedback. The CNN model outperformed the SVM model with an F-measure of 0.78, 0.73, and 0.71 for the 5th week, 7th week, and 9th week of a semester. Similarly, the CNN model was used to evaluate lecturer effectiveness based on student feedback to a questionnaire by Sutoyo et al.~\cite{Sutoyo2021}. The model was able to achieve accuracy, precision, recall, and F1-Score of 87.95\%, 87\%, 78\%, and 81\%, respectively. Deep learning models can be further enhanced by adding attention layers to identify sadness influence of words on emotion. Sangeetha et al.~\cite{Sangeetha2020} proposed a multi-head attention fusion model for sentiment analysis of student feedback. The input sequences of sentences from feedback are processed in parallel across the multi-head attention layer with word and context embeddings like Glove and Cove. The outputs of the two multi-head attention layers with the embeddings are passed to the deep learning model LSTM. The dropouts of these layers are regulated to improve the accuracy of the model. The proposed fusion model was able to classify three sentiment orientations positive, negative, and neutral more accurately when compared to individual multi-head attention and LSTM. 

\subsubsection{Transformers}
Dyulicheva et al.~\cite{dyulicheva2022learning} proposed a bidirectional encoder representation from transformers (BERT) model for sentiment detection of students, identifying top words describing positive and negative polarities. The authors used K-Means clustering and cosine similarity on 300 MOOCs titles from Udemy to extract 14 clusters and top words in each cluster. The BERT model was used to investigate the relationship between student-teacher, student-course, and description of issues while learning. The results show more negative sentiment towards courses when compared to teachers. Similarly, Li et al.~\cite{Li2019} analysed the sentiment of learning comments MOOCs using a shallow BERT-CNN model. The authors took advantage of deep learning not to depend on feature engineering and ensemble with the BERT model. The proposed BERT-CNN with a self-attention mechanism performed nearly equal to the traditional BERT model even after reducing the number of parameters to half. The proposed shallow BERT-CNN model with 6 layers outperformed all other lexicon-based methods, BERT variants with just 61 million parameters with an accuracy, F1-Score(Positive class), and F1-Score(Negative class) of 92.8\&, 95.2\%, and 81.3\% respectively.

\section{Sentiment Analysis Impact on Education}\label{impact}
The research community has widely adopted sentiment analysis to extract student opinions toward teaching and learning practices in the education sector~\cite{Han2019}. The opinion mining mechanism can be adapted to different educational domain applications such as understanding student engagement pedagogy, limitations of existing educational infrastructure, course and teaching evaluation, decision-making on educational policies, and so on. These applications can be achieved by understanding student feedback on a course in a semester, and their participation in online forums.

The role of sentiment analysis in the education domain is more toward student feedback analysis and enhancing learning-teaching practices~\cite{Zhou2020}. Sentiment analysis can also be adapted to personalise learning for individual students by understanding their issues, aspect-based, and entity-level opinions~\cite{DAniello2022}. Text categorisation is to assign predefined topics in a comment. From a sentiment perspective, text can be categorised using sentence-level analysis. Emotional analysis summarises student feedback on a course into graded sentiment analysis with scales like very positive, positive, neutral, negative, and very negative~\cite{Sharma2020}.  

\subsection{Learning and Teaching Systems Evaluation}
Teacher evaluation is part of education systems evaluations and it provides constructive feedback for teachers' professional growth. Tertiary education institutions conduct student feedback surveys on courses taught in that semester. The feedback has both qualitative and quantitative feedback on the courses. An ensemble learning of five machine learning algorithms such as Naive Bayes, logistic regression, support vector machine, decision tree, and random forest classifiers based on the majority voting principle was proposed by Lalata et al.~\cite{Lalata2019}. The sentiment of student comments collected from each semester was manually labelled into positive, negative, and neutral comments. The authors performed individual classification and ensembled the classifier output. The results show all model ensembling has achieved accuracy, F1-score, and recall of 90.32\%, 93.80\%, and 90.86\%. Roaring et al.~\cite{Roaring2022} analysed the relationship between the numerical rating of faculty performance and the actual opinions, feelings, and observations using text analytics. Association rule mining was used for pattern recognition based on rule-based metrics support, confidence, and lift. The faculty rating on a scale of 1-5 was evaluated in the study. Negative words were used in their feedback by students to attribute faculty with a rating of 1. Similarly, neutral words for faculty with a rating of 2, and positive words for faculty with a rating of 3. The feedback had positive observations and opinions for faculty with a rating of 4 and 5. Pramod et al.~\cite{Pramod2022} proposed a two-fold faculty evaluation system based on machine learning and text analytics. A structured questionnaire was used to record student and faculty characteristics, and qualitative and quantitative feedback from post-graduate students about their faculty. The qualitative feedback was analysed using sentiment analysis and converted the text feedback to polarity scores. The polarity of the words and sentences in the qualitative feedback varied across faculty and the time span of feedback. The engaging faculty members used case studies, imparting practical experiments, and real-time examples and they had high emotional engagement from students. The polarity scores of the qualitative feedback were combined with quantitative feedback to predict faculty effectiveness using ten different machine learning algorithms. Out of all, the random forest model achieved high accuracy, precision, and area under the curve of  98.87\%, 97.71\%, and 97.32\% respectively.

Massive open online courses (MOOCs) deliver the learning content to participants breaking barriers like age, gender, race or geography. Onan~\cite{ONAN2020} evaluated the predictive performances of conventional machine learning, ensemble learning methods, and deep learning algorithms based on 66,000 student reviews for MOOCs. The conventional supervised methods and ensemble techniques used three-term weighting techniques: term presence, term frequency, and TF-IDF. Three word-embedding techniques word2vec, fastText, and GloVe were used in deep learning models. Out of all, the deep learning model LSTM with the GloVe word-embedding technique had the highest classification accuracy of 95.80\%.

\subsection{Enhance Pedagogical Concepts}
Analysing student feedback each semester reveals issues in the pedagogical structure of teaching and learning practices. Student engagement is one of the key pedagogical concepts in education. Yan et al.~\cite{Yan2019} evaluated learner engagement in Teach-Outs, which aims to drive meaningful discussions around topics of social urgency without any formal assessment. The authors used opinion-mining techniques including sentiment analysis, topic modelling, and similarity analysis to investigate the topics discussed by learners in the online forum. Latent Dirichlet Allocation (LDA)~\cite{blei2003latent}, a probabilistic topic modelling model technique, was used to extract topics from forum posts and the positive and negative emotions of each post were detected using Linguistic Inquiry and Word Count (LIWC)~\cite{pennebaker2015development}. Li et al.~\cite{Li2022} derived two distinct categories: knowledge-seeking and skill-seeking courses in MOOCs pedagogy by applying topical ontology of keywords and sentiment techniques to learners' comments on top-rated courses on Coursera.org. Knowledge-seeking courses stress learning concepts that provide insights, assessed with quizzes and final exams and they are driven by course design and materials. Skill-seeking courses require learning techniques through labs, projects, or assignments and they are driven by instructors. Sentence-level sentiment analysis was used for sentiment classification to detect the polarity of a sentence based on identified topics and to perform a fine-grained analysis of opposite sentiments on different topics. A learner review is chunked into multiple sentences and extracted the topic and sentiment for each sentence. To deliver pedagogical cognitive and affective skills, conversational agents based on reinforcement learning (RL) and sentiment analysis were proposed by Feidakis~\cite{feidakis2019building}. In that study, the learning task is expressed in a goal-oriented way to RL agents. The agents learn how to dialogue with students based on supervised learning of annotated datasets and the learning is through a trial-and-error method based on the incoming rewards for each action.

\begin{landscape}
\begin{table}[]
\centering
\caption{Sentiment Analysis tools in Educational Domain}
\label{tab:SA_Tools}
\resizebox{\columnwidth}{!}{
\begin{tabular}{@{}ccccc@{}}
\toprule
\textbf{Tool} &
  \textbf{Application} &
  \textbf{Algorithms} &
  \textbf{Performance} &
  \textbf{Reference} \\ \midrule
\begin{tabular}[c]{@{}c@{}}Suggestion \\ Extraction Tool\end{tabular} &
  \begin{tabular}[c]{@{}c@{}}Topic extraction\\ sentiment analysis\\ suggestion extraction \\ to improve instructor teaching\end{tabular} &
  \begin{tabular}[c]{@{}c@{}}GLM, SVM, \\ CIT, DT\end{tabular} &
  \begin{tabular}[c]{@{}c@{}}DT with precision 0.802, \\ recall 0.775, and F-score 0.781\end{tabular} &
  \cite{Gottipati2018, Pyasi2018} \\ \midrule
\begin{tabular}[c]{@{}c@{}}Student Textual Feedback \\ Analysis Tool\end{tabular} &
  Topic extraction, sentiment analysis &
  \begin{tabular}[c]{@{}c@{}}LDA, TextBlob, \\ Polarity Analyser\end{tabular} &
  \begin{tabular}[c]{@{}c@{}}TextBlob improved version \\ with F-Score 0.92\end{tabular} &
  \cite{gottipati2018latent} \\ \midrule
\begin{tabular}[c]{@{}c@{}}Automated Discussion \\ Analysis (ADA) framework\end{tabular} &
  \begin{tabular}[c]{@{}c@{}}Question Answering Systems, \\ Recommendation Systems, \\ Summarisation Systems, \\ Opinion Mining Systems,\\ Information retrieval systems\end{tabular} &
  Clustering techniques &
   &
  \cite{GOTTIPATI2020} \\ \midrule
\begin{tabular}[c]{@{}c@{}}SETSUM: Summarization and \\ Visualisation of Student \\ Evaluations of Teaching\end{tabular} &
  \begin{tabular}[c]{@{}c@{}}Aspect Annotation, \\ Sentiment Analysis, \\ Extractive Summarisation\end{tabular} &
  \begin{tabular}[c]{@{}c@{}}LexRank, LDA, \\ manual sentiment annotation\end{tabular} & 
   &\cite{hu2022setsum}
   \\ \midrule
Palaute tool &
  \begin{tabular}[c]{@{}c@{}}Topic modelling, \\ emotional analysis\end{tabular} &
  LDA, Syuzhet &
   &
  \cite{Grnberg2020, Gronberg2021} \\ \midrule
\begin{tabular}[c]{@{}c@{}}AI-enabled adaptive \\ learning system\end{tabular} &
  \begin{tabular}[c]{@{}c@{}}Generate personalised quizzes \\ for each learner\end{tabular} &
  \begin{tabular}[c]{@{}c@{}}Reinforcement Learning, \\ LDA\end{tabular} &
  Net Promoter Score 38 &
  \cite{tan2022ai} \\ \bottomrule
\end{tabular}
}
\end{table}
\end{landscape}

\subsection{Decision Making}
Educational data mining assists educational institutions in measuring the teaching and learning process and improving their student recruitment and retention policies. Hussian et al.~\cite{Hussain2022} proposed a decision support system based on a multi-layered Aspect2Labels (A2L) approach. It is a three-layered topic modeling approach, the first layer extracts high-level aspects preserving semantic information, the second layer extracts low-level aspects associated with high-level aspects and the third layer was used for sentiment orientation. The authors annotated unlabelled student comments using the proposed A2L approach and achieved 91.3\% accuracy. Two techniques Variable Stopwords Filtering (VSF) and Variable Global Feature Selection Scheme (VGFSS) proposed to improve the performance of the classifier. The authors used five machine learning models for sentiments analysis and Support Vector Machine with the linear kernel (SVM) and Artificial Neural Network (ANN) algorithms performed better than other models with an accuracy of 97\% and 93\% respectively. 

\subsection{Assessment Evaluation}
Distinctive understanding of a concept and the polarity of the words used by students shape the quality and presentation of a course assignment or assessment~\cite{Janda2019}. In ~\cite{Janda2019}, the authors proposed to use the semantic similarity of sentences of text along with their sentiments, and sentence syntax for automated essay evaluation. For example, if a student writing an argumentative and persuasive essay, he/she has to prove their view on the topic, check the tone, and the way of textual sentiment expression. The authors used Vader, a lexical approach to extract the polarity for each sentiment of the words used in the essay. Gkontzis~\cite{Gkontzis2020} analysed the activity, polarity, and emotions of students and tutors to predict students' grades. The authors used six moodle datasets of postgraduate programs at the Hellenic Open University. The results show that extracting the polarity and emotions of students can provide a better prediction of students' performance. Furthermore, variables related to tutors also contribute to student performance. The study has compared actual and predicted grades and investigated which students had used third-party services to do their assignments. The research works outlined in Tab.~\ref{tab:sa_impact} present the AI algorithms that can be adapted to education for its applications.

\begin{table}[!h]
\scriptsize
\caption{SA impact on education}
\label{tab:sa_impact}
\resizebox{\textwidth}{!}{%
\begin{tabular}{@{}lll@{}}
\toprule
\textbf{\begin{tabular}[c]{@{}l@{}}Educational \\ Applications\end{tabular}} &
  \textbf{References} &
  \textbf{Algorithms} \\ \midrule
\multicolumn{1}{l}{\multirow{4}{*}{\begin{tabular}[c]{@{}l@{}}Learning and Teaching \\ Systems Evaluation\end{tabular}}} &
  \multicolumn{1}{l}{Lalata et al.~\cite{Lalata2019}} &
  \multicolumn{1}{l}{Ensemble learning} \\ \cmidrule(l){2-3} 
\multicolumn{1}{l}{} &
  \multicolumn{1}{l}{Roaring et al.~\cite{Roaring2022}} &
  \multicolumn{1}{l}{Association rule mining} \\ \cmidrule(l){2-3} 
\multicolumn{1}{l}{} &
  \multicolumn{1}{l}{Pramod et al.~\cite{Pramod2022}} &
  \multicolumn{1}{l}{RF} \\ \cmidrule(l){2-3} 
\multicolumn{1}{l}{} &
  \multicolumn{1}{l}{Onan~\cite{ONAN2020}} &
  \multicolumn{1}{l}{\begin{tabular}[c]{@{}l@{}}word2vec, fastText, \\ GloVe, LSTM\end{tabular}} \\ \midrule
\multicolumn{1}{l}{\multirow{3}{*}{\begin{tabular}[c]{@{}l@{}}Enhance Pedagogical \\ Concepts\end{tabular}}} &
  \multicolumn{1}{l}{Yan et al.~\cite{Yan2019}} &
  \multicolumn{1}{l}{LDA, LIWC} \\ \cmidrule(l){2-3} 
\multicolumn{1}{l}{} &
  \multicolumn{1}{l}{Li et al.~\cite{Li2022}} &
  \multicolumn{1}{l}{Topic Ontology} \\ \cmidrule(l){2-3} 
\multicolumn{1}{l}{} &
  \multicolumn{1}{l}{Feidakis~\cite{feidakis2019building}} &
  \multicolumn{1}{l}{reinforcement learning} \\ \midrule
\multicolumn{1}{l}{Decision Making} &
  \multicolumn{1}{l}{Hussian et al.~\cite{Hussain2022}} &
  \multicolumn{1}{l}{VSF, VGFSS, SVM, ANN} \\ \midrule
\multicolumn{1}{l}{\multirow{2}{*}{Assessment Evaluation}} &
  \multicolumn{1}{l}{Janda et al.~\cite{Janda2019}} &
  \multicolumn{1}{l}{Vader} \\ \cmidrule(l){2-3} 
\multicolumn{1}{l}{} &
  Gkontzis~\cite{Gkontzis2020} &
  \begin{tabular}[c]{@{}l@{}}Additive regression,\\ RF, SVN, KNN\end{tabular} \\ \bottomrule
\end{tabular}}
\end{table}

\subsection{Sentiment Analysis Tools in Education}
Educational institutions have been working on building student feedback mining tools to analyse their students' opinions on existing learning infrastructures, teachers, and course contents. With the analysis, the institutions' intuition is to evaluate their teaching practices and learning management systems based on suggestions in student feedback. A few of those research and application works are enclosed in Table~\ref{tab:SA_Tools}. AI algorithms are explicitly used in most of the works. Advanced AI methodologies like reinforcement learning would enable the education sector to personalise student learning experiences and support them.

\section{Challenges}\label{challenges}
\subsection{Negation Handling}
In sentiment analysis, the textual feedback received from students or instructors gets preprocessed to remove stopwords and prepare for sentiment labelling. The stopwords are low-level information that doesn't add much value to sentiment analysis. By eliminating the stopwords, the analysis can focus on important information such as noun, verb, and adjective words of the textual feedback. However, the negation words such as never, not, none, don't, and so on in negative statements are preprocessed and these auxiliary verbs and adverbs are removed. For example, the negative statement "Lecturer never discuss practical knowledge concepts" passed through preprocessing step would remove "never" and the text analytic models get trained on inverse sentiment. 

In the Aspect2Labels framework proposed by Hussain et al.~\cite{Hussain2022}, the authors reversed the opinion score to -1 whenever negative words not, never, nothing, neither, or none are encountered in student feedback. The authors also dealt with blind negation words such as need, needed, require, and required by assigning a negative polarity score to the sentence. Singh et al.~\cite{Singh2021} proposed focusing on cue and scope could find the polarity shift in a sentence. A deep learning model LSTM was proposed to learn negation features of pre-annotated ConanDoyle corpus with negation information. The LSTM model identifies negation cues and BiLSTM extracts the relationship between the extracted cues and other words. The proposed approach was compared with traditional machine learning methods SVM, Hidden Markov Model, and CRF models. BiLSTM model outperformed other models with an F-Score of 93.34\% in negation handling tasks. Similarly, Gupta et al.~\cite{Gupta2021} proposed a feature-based Twitter sentiment analysis (TSA) framework based on cues and the scope of sentences for negation handling. The research work is an improvement to the work proposed by Mohammed et al.~\cite{mohammed2013} in negation handling in which the authors argue negation presence not always led to the negative polarity of a sentence.

\subsection{Opinion Spam Detection}
In the education domain, opinion spam detection is critical as students might use third-party services for assessment items and write fake feedback to enrolled courses at the end of the semester. The opinion spam detection concept has been widely explored in product reviews where people get incentives for writing fake reviews~\cite{Yuan2019}.

Saumya et al.~\cite{Saumya2018} discussed three challenges in the supervised approach to detect spam comments. They are manual labelling of comments into spam and non-spam, data imbalance problems due to scarcity of spam comments, and expensive computation to retrieve similarity among the spam comments. The authors proposed a system to overcome the labelling challenge and automate spam comments annotation. In that study, RF, SVM and Gradient Boosting classifiers were used to classify comments into spam or non-spam classes. Two oversampling techniques Synthetic Minority Oversampling Technique (SMOTE)~\cite{Chawla2002} and Adaptive Synthetic (ADASYN)~\cite{HaiboHe2008} to balance the datasets with the same number of spam and non-spam records. Deep learning methods were adopted for opinion spam detection by Bathla et al.~\cite{Bathla2021}. Convolutional Neural Networks (CNN) model was used to classify the spam comments with the GloVe embedding technique. The results show that the deep learning method outperformed traditional machine learning methods such as SVM, KNN, Logistic Regression, Naive Bayes, MLP and other deep learning methods LSTM.

\subsection{Multi-Polarity}
Multi-polarity refers to several contrasting ideas and emotions towards a product or service. In the education domain, student feedback might possess such multi-polarity challenges in which students might have positive emotions towards one aspect of a course and negative opinions on other aspects. In such situations, the sentiment analysis of the student feedback would be challenging as it might be ambiguous or misleading for sentiment annotation techniques to label the text. 

Wei et al.~\cite{Wei2020} proposed a BiLSTM model with multi-polarity orthogonal attention for implicit sentiment analysis. Dictionary-based sentiment knowledge such as neutral, positive, and negative embeddings are trained in the polarity layer with an orthogonal attention mechanism. Wang et al.~\cite{Wang2018} present a semantic analysis model (SMA) framework to track the emotional tendencies of students based on comments, forum posts, and homework completion on the MOOC platforms. In that study, the SMA framework detects learners' multi-polarity sentiments by adopting the NRC lexicon for sentiment quantification.  

\begin{table}[]
\scriptsize{}
\caption{Sentiment Analysis Challenges in Education}
\label{tab:SA_Challenges}
\resizebox{\textwidth}{!}{%
\begin{tabular}{@{}lll@{}}
\toprule
\textbf{Challenges}       & \textbf{Trends}                          & \textbf{References}                                           \\ \midrule
\multicolumn{1}{l}{Negation Handling} &
  \multicolumn{1}{l}{Negative polarity scoring} &
  \multicolumn{1}{l}{~\cite{Hussain2022, Singh2021, Gupta2021,mohammed2013}} \\ \midrule
\multicolumn{1}{l}{Opinion Spam Detection} &
  \multicolumn{1}{l}{SMOTE, ADASYN, Deep Learning} &
  \multicolumn{1}{l}{\cite{Yuan2019,Saumya2018,Bathla2021}} \\ \midrule
\multicolumn{1}{l}{Multi-Polarity} &
  \multicolumn{1}{l}{Dictionary-based sentiment knowledge} &
  \multicolumn{1}{l}{\cite{Wei2020, Wang2018}} \\ \midrule
 Polysemous Words & ACWE, Word Sense Disambiguation & \cite{Logacheva2020, Li2021, Jia2021, Yenicelik2020} \\ \bottomrule
\end{tabular}}
\end{table}

\subsection{Polysemous Words}
Polysemous refers to a word that has several meanings. This would cause word sense disambiguation (WSD)~\cite{Logacheva2020} to AI models and sentiment annotations while sentiment analysis. The word embedding concept in NLP represents each word by a vector to capture the semantic correlation between words. The research community tried to overcome the challenge by generating a small number of vectors for each word but it is difficult to estimate the number of meanings for each word. 

Li et al.~\cite{Li2021} proposed a novel adaptive cross-contextual word embedding (ACWE) method to capture polysemous words in different contexts using topic modelling by defining a latent interpretable semantic space. A global word embedding was obtained based on an unsupervised cross-contextual probabilistic word embedding model and represents each word in the unified latent semantic space. An adaptive cross-contextual word embedding process is then defined to learn local word embedding for each polysemous word in different contexts. Jia et al.~\cite{Jia2021} proposed an effective topical word embedding (TWE)-based WSD method to generate topical word vectors for each word under each topic. The authors used LDA to extract high-quality topic models, and then generate contextual vectors generated for an ambiguous word to exploit the context. Based on the topic models and contextual vectors, topical word vectors were generated. Yenicelik et al.~\cite{Yenicelik2020} used the BERT model to produce embedding vectors using quantitative analysis of linear separability and cluster organisation. The authors trained a simple linear classifier on top of the BERT embeddings to predict the semantic class and report accuracy. The challenges and trends in adopting sentiment analysis to education have been discussed and outlined in the research works in Tab.~\ref{tab:SA_Challenges}.

\section{Future Directions of SA in Education}\label{future}
In this study, a comprehensive review of sentiment analysis in the educational context is conducted. The impact of sentiment analysis in the educational domain and the challenges in adopting sentiment analysis pave a path to future research directions. Sentiment analysis can assist educational institutions to evaluate their learning and teaching practices by extracting sentiment from student opinions. Advancements in transformers and deep learning techniques transform sentiment analysis applications in various fields. 

The future direction should be focused on education-based sentiment annotation techniques. In current trends, the research community has adopted existing unsupervised sentiment annotation techniques such as corpus-based and lexicon-based approaches. However, there will be a huge semantic difference in student feedback on the education context and the language or words used in the unsupervised annotation techniques. Focusing on education-based unsupervised annotation techniques can overcome lexical ambiguity, and data labelling challenges and understand the core factors of educational context. To build an education-based annotation, a corpus of education documents has to be labelled manually in a dictionary-based approach. Based on the education-based annotation technique, the reinforcement learning process can be enhanced to provide improved support for goal-oriented learning tasks discussed in \cite{feidakis2019building}. 

Topic ontology discussed in \cite{Li2022, Tao2008} can be further explored to build an educational knowledge base and focus on defining relationships among multimodal data collected in student feedback. With knowledge-based systems enhance the educational data mining process and overcome challenges such as sarcasm, negation handling, and ambiguity. Fine-tuning of aspect-based and entity sentiment analysis can be achieved by annotating entities and aspects that need to be emphasised in an educational domain.

\section{Conclusions} \label{Conclusions}
In this review article, a comprehensive impact of sentiment analysis and opinion mining on educational data and challenges are explored. Though, sentiment analysis is the simple detection of the sentiment orientation of students towards course content or teaching, fine-grained analysis in education would require other NLP techniques like topic modelling. To analyse a large number of students' opinions, the usage of AI methodologies such as machine learning, deep learning, and transformers is inevitable. 

This study reviews the role of sentiment analysis in the enhancement of educational procedures such as decision-making, pedagogical concepts, and evaluation systems. The pandemic situation expedites the process of online education all around the world and enlarges the MOOCs industry. This review article covered studies related to both offline and online modes of education. The research community has been working on aspect-level sentiment analysis for a fine-grained analysis at the topic level and to extract the sentiments of students. Dictionary and corpus-based techniques are unsupervised sentiment annotation techniques discussed in the study. Throughout the review, most of the researchers used AI methods like traditional machine learning, deep learning, or BERT models to analyse, predict and classify the opinions of a large group of students. Yet, there are challenges in sentiment analysis that need attention to improvise the existing student feedback and mining techniques.



\bibliographystyle{elsarticle-num}
\bibliography{biblio}
\end{document}